\documentclass[10pt,english,english,10pt,journal,english,compsoc]{IEEEtran}
\usepackage[latin9]{inputenc}
\usepackage{color}
\usepackage{babel}
\usepackage{float}
\usepackage{textcomp}
\usepackage{amsmath}
\usepackage{amssymb}
\usepackage{stmaryrd}
\usepackage{graphicx}

\providecommand{\tabularnewline}{\\}
\floatstyle{ruled}
\newfloat{algorithm}{tbp}{loa}
\providecommand{\algorithmname}{Algorithm}
\floatname{algorithm}{\protect\algorithmname}

\providecommand{\tabularnewline}{\\}
\floatstyle{ruled}
\newfloat{algorithm}{tbp}{loa}
\providecommand{\algorithmname}{Algorithm}
\floatname{algorithm}{\protect\algorithmname}

\usepackage{babel}

\usepackage{url}

\usepackage{array}
\usepackage{multirow}

\usepackage[T1]{fontenc}
\usepackage[latin9]{inputenc}
\usepackage{color}
\usepackage{booktabs}
\usepackage{amstext}
\usepackage{amssymb}
\usepackage{graphicx}

\makeatletter

\pdfpageheight\paperheight
\pdfpagewidth\paperwidth

\let\SF@@footnote\footnote
\def\footnote{\ifx\protect\@typeset@protect
    \expandafter\SF@@footnote
  \else
    \expandafter\SF@gobble@opt
  \fi
}
\expandafter\def\csname SF@gobble@opt \endcsname{\@ifnextchar[
  \SF@gobble@twobracket
  \@gobble
}
\edef\SF@gobble@opt{\noexpand\protect
  \expandafter\noexpand\csname SF@gobble@opt \endcsname}
\def\SF@gobble@twobracket[#1]#2{}
\providecommand{\tabularnewline}{\\}

\usepackage{algorithm}
\usepackage{algorithmic}

\usepackage{amssymb}

\makeatother

\begin{document}

\title{Learning Classifiers from Synthetic Data Using a Multichannel Autoencoder}

\author{Xi~Zhang, Yanwei~Fu, Andi~Zang, Leonid~Sigal, Gady~Agam 
\IEEEcompsocitemizethanks{
\IEEEcompsocthanksitem Xi Zhang, Andi Zang, and Gady Agam are with the Illinois Institute of Technology Chicago, IL 60616.\protect\\ 
Email: \{xzhang22,zang\}@hawk.iit.edu, and agam@iit.edu.
\IEEEcompsocthanksitem Yanwei~Fu, and Leonid~Sigal  are with  Disney Research, Pittburgh, PA, 15213. \protect\\ 
 Email: \{yanwei.fu, lsigal\}@disneyresearch.com}
\thanks{}

}

\IEEEcompsoctitleabstractindextext{ 
\begin{abstract}
We propose a method for using synthetic data to help learning classifiers. Synthetic data, even is generated  based on real data, normally results in a shift from the distribution of real data in feature space. To bridge the gap between the real and synthetic data, and jointly learn from synthetic and real data, this paper proposes a Multichannel Autoencoder(MCAE). We show that by suing MCAE, it is possible to learn a better feature representation for classification. To evaluate the proposed approach, we conduct experiments on two types of datasets. Experimental results on two datasets validate the efficiency of our MCAE model and our methodology of generating synthetic data.

\end{abstract}
}

\maketitle

\section{Introduction}
Large and balanced datasets are normally crucial for learning classifiers. 
In real-world scenarios, however, one always struggles to find adequate
amounts of labeled data. Even with the help of crowdsourcing, e.g.,
Amazon Mechanical Turk (AMT), it is often difficult to collect a large 
quantity of labeled instances with high quality that is necessary for 
training a classifier for a real-world problem. 
In terms of quantity, it has been shown that the amount of
available training data, per object class, roughly follows a Zipf distribution \cite{torralba2011app_share}. That means a small number of 
object classes account for most of the available training instances.
In terms of quality, some domains, such as the analysis of satellite images
(e.g. the comet images from Rosetta), require extensive and detailed
expert user annotation \cite{remotesensing2013,ZX:14b}. Large volume
of LiDAR point cloud data have to be labeled before they can be used
to train some classifiers \cite{ZX:14}. Such labeling process usually is very time consuming and requires expert-level labeling efforts or expensive equipments. Practically only a very limited portion of the data points
can be obtained.

\begin{figure}
\centerline{ %
\begin{tabular}{cc}
\resizebox{0.23\textwidth}{!}{\rotatebox{0}{ \includegraphics{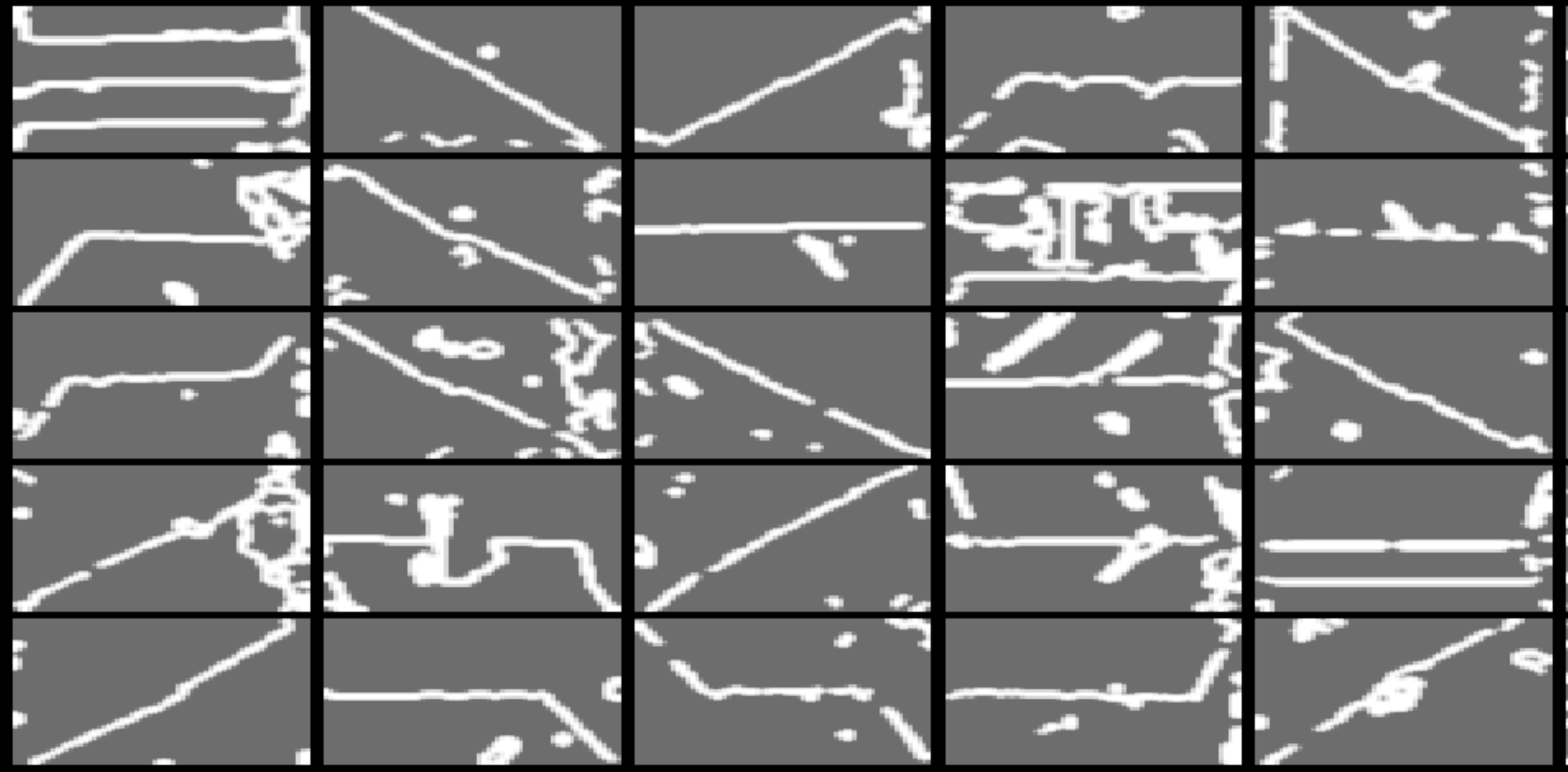}}}  
& 
\resizebox{0.23\textwidth}{!}{\rotatebox{0}{ \includegraphics{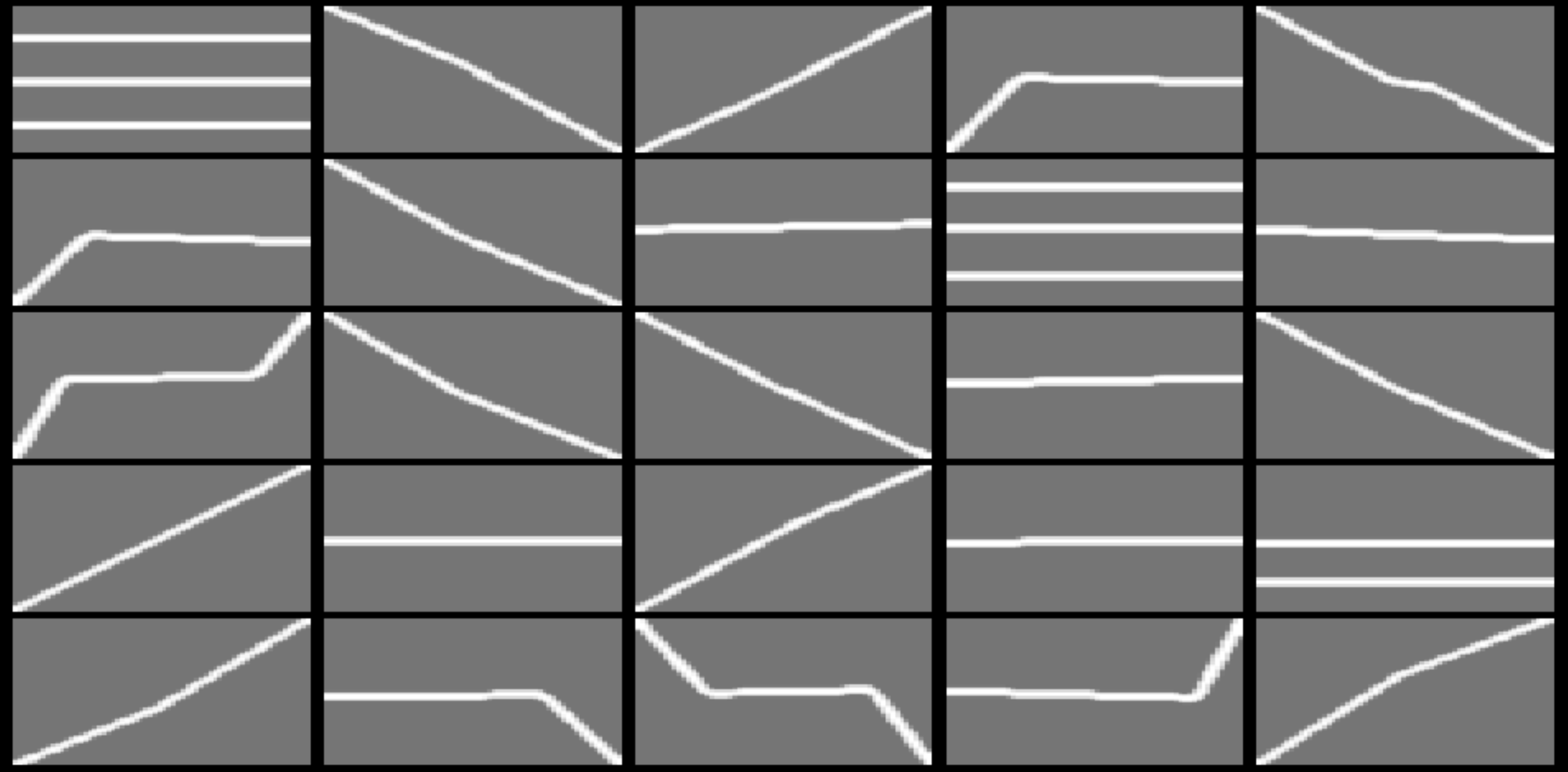}}} 
\tabularnewline
(a) Real roof edges & (b) Synthetic roof edges 
\tabularnewline
\end{tabular}} \protect\caption{Examples of (a) real roof edge vs. corresponding (b) synthetic roof
edge images. The synthetic data is generated by the algorithms in Sec.
4. The examples are randomly drawn from the SRC dataset.}

\label{fig: roof real and synI} 
\end{figure}

To solve the problem of lacking enough training samples,
attributes \cite{lampert2009zeroshot_dat,palatucci2009zero_shot,farhadi2009attrib_describe}
have been introduced to transfer the knowledge held by majority classes to instances in minority classes. 
Nevertheless, for certain tasks, such shared
attributes~\cite{fu2012attribsocial,Dhar2011cvpr,liu2011action_attrib,datta2011face_attrib,Yucatergorylevel}
may simply be unavailable or nontrivial to define.
In contrast, rather than using such a `learning to learn' \cite{Thrun96learningto} framework, humans can
generalize and associate the similar patterns from images. This
ability inspires us to circumvent the problem of lacking enough training data and solve it from a different
angle: utilizing the synthetic data (e.g. the synthetic roof edges in Fig.
\ref{fig: roof real and synI}) associated with real data (e.g.
real roof edges in Fig. \ref{fig: roof real and synI}) in order to learn a better classifier. 

The idea of associating synthetic data with real data has a long
history and is associated with the development of cognitive psychology, 
artificial intelligence, and computer vision. For example, cognitive psychology
studied a case that an infant learns to understand and imitate a facial
expression from parents' examples. In the computing domain, exemplar SVM \cite{examplarSVM}
tries to associate images with training exemplars. Different
from these previous works, we create synthetic images to associate them with 
the real images whilst previous works associate `new' real
data with `old' real data. By contrast, our approach is a 'free lunch'
in the sense that the proposed approach does not need any human annotation of real data, 
thus we could easily amplify the dataset used in training.

\begin{figure}
\centering
\begin{tabular}{cc}
\includegraphics[scale=0.29]{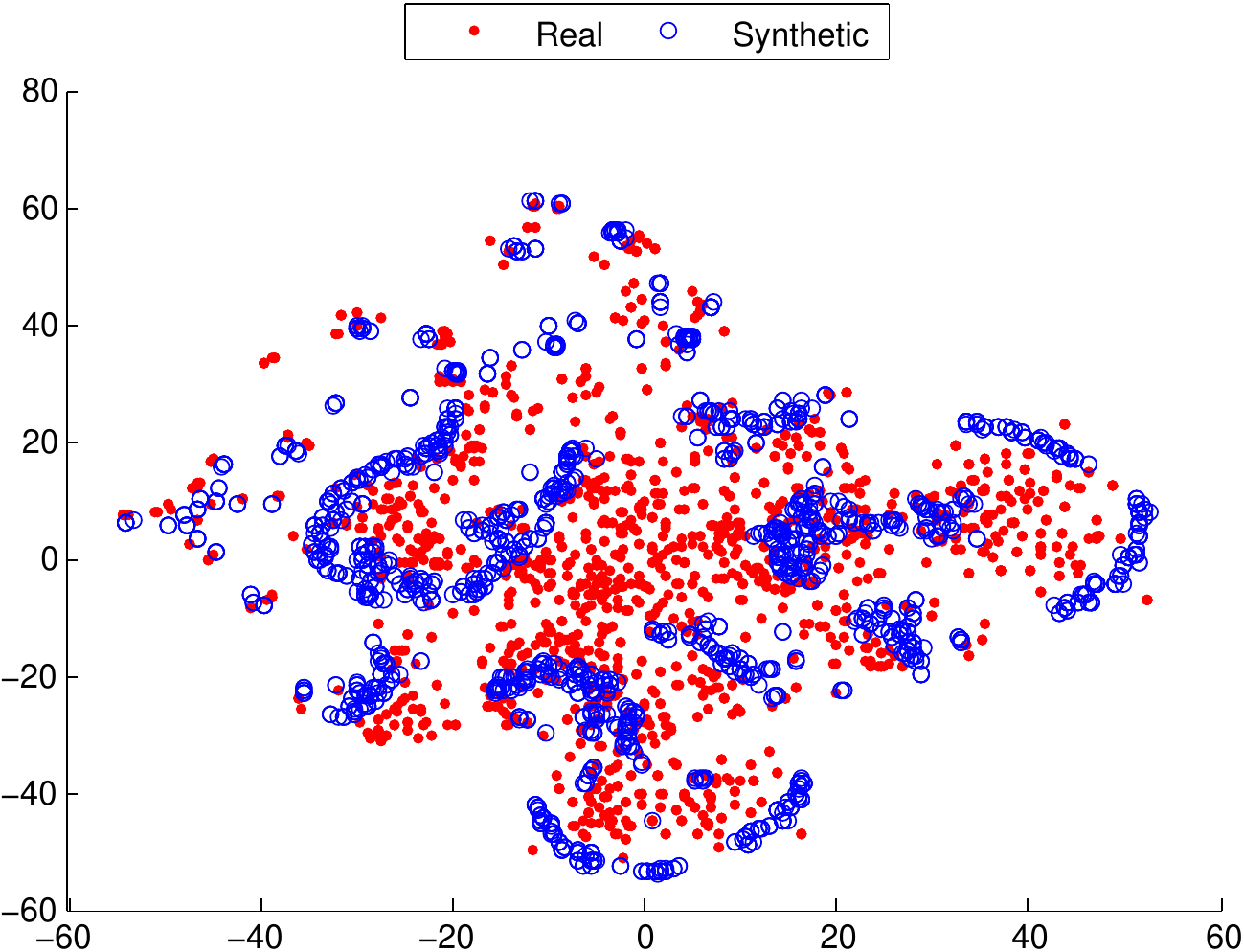} 
& 
\includegraphics[scale=0.29]{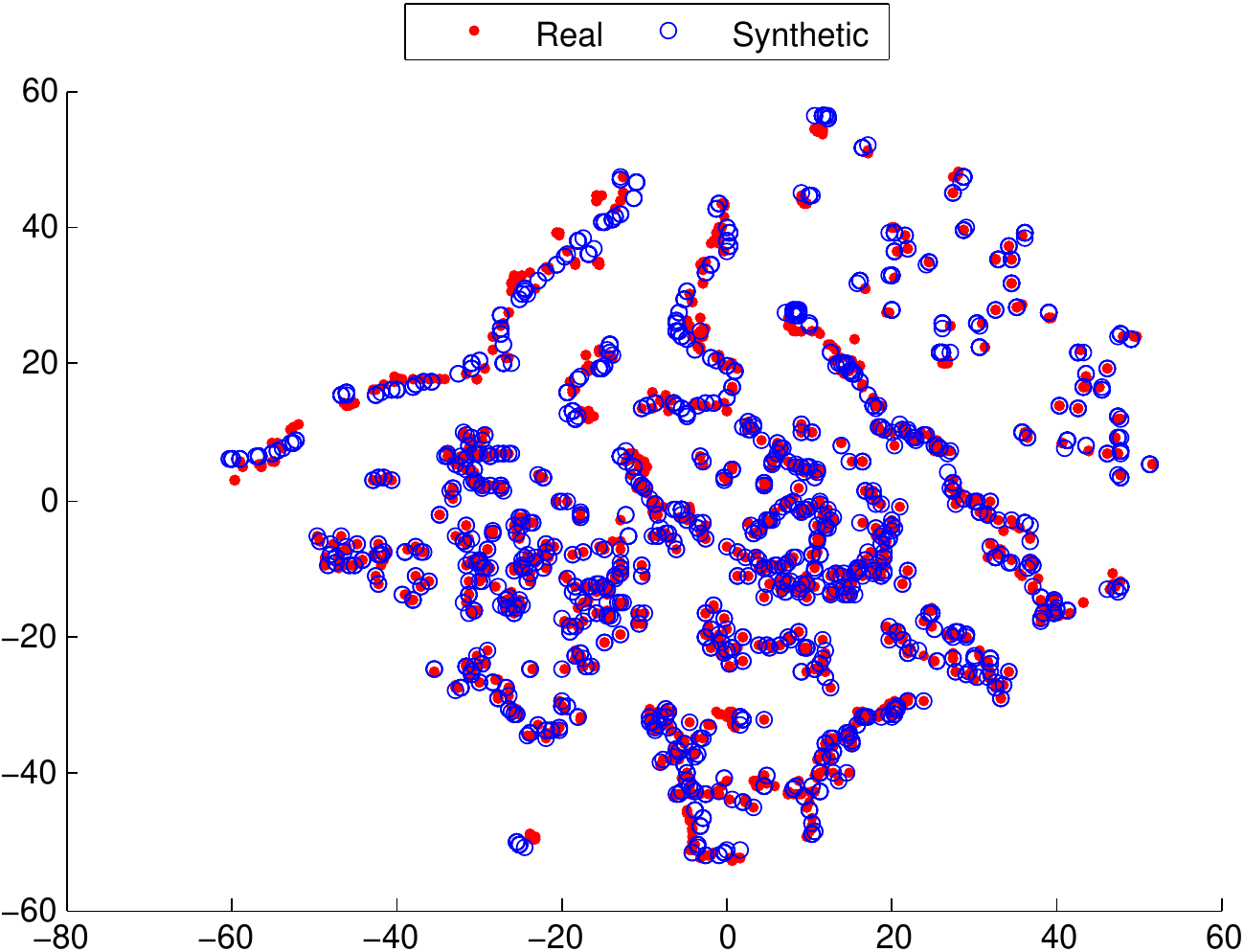}\tabularnewline
(a) & (b)\tabularnewline
\end{tabular}\protect\caption{t-SNE visulization of synthetic gap using the data from SRC dataset. (a) synthetic gap of real and
synthetic data; (b) MCAE bridges the synthetic gap.}

\label{fig: tsne_vis} 
\end{figure}

Learning a classifier from synthetic data is unfortunately extremely
challenging due to the following reasons. Firstly, the feature distribution
of synthetic data generated will shift away from that of real data.
Such distribution shift is termed \emph{synthetic gap} and illustrated
in Fig \ref{fig: tsne_vis}. The synthetic gap is
a major obstacle in using synthetic data to help learning classifiers,
since synthetic data may fail to simulate the potential useful patterns
of real data for training classifiers. To our knowledge, this synthetic
gap problem has never been formally identified nor addressed in the literature.
Secondly, since practically a small amount of labeled images may be
available, it is necessary to jointly learn from synthetic and real
data. The learning process must be automatically leveraged between
synthetic data and real data. 


\textcolor{red}{}

To better learn a classifier from synthetic data, we propose a novel
framework --Multichannel Autoencoder (MCAE) which is an extension of sparse autoencoder. The training step of
MCAE is a process of bridging the synthetic gap between the real and
the synthetic data by learning the mapping from (1) synthetic
to real data and (2) real to real data. Critically, such mapping try to keep the real data 
while enforce MCAE to learn a transfer from the synthetic data to the real data. 
We thus can generate more synthetic data which will simulate the real data when the learned mapping is applied to them.


To facilitate the study on satellite image analysis,
we introduce a new benchmark satellite roof classification (SRC) image dataset.
The SRC dataset needs expert-level
labeling and has unique challenges, such as satellite image blurring, 
building shadows, and extremely imbalanced roof class
instances. To demonstrate the generality of the proposed approach, we use an additional
handwritten digit dataset from the UCI machine learning repository \cite{Bache+Lichman:2013}.
In both datasets, synthetic data is generated using a parametric model of derived from real data that roughly mimics real data in terms of appearances and basic structure. Experimental results using these datasets demonstrate that better classification results can
be obtained by training a classifier using the synthetic data when used
by the proposed approach.

We thus highlight three contributions in this paper: (1) To the best
of our knowledge, this is the first attempt to address the problem of synthetic gap, by solving which we demonstrate that the synthetic data could be used to improve the performance of classifiers. (2) We propose a Multichannel Autoencoder (MCAE) model to bridge
the synthetic gap and jointly learn from both real and synthetic data.
(3) Also, a novel benchmark dataset -- Satellite Roof Classification (SRC) is
introduced to the vision community. Such dataset is of expert-level
label annotations as well as great challenges for satellite image
analysis.


\section{Related Work}

\textbf{3D image analysis. }Synthetic data has been used for several
3D image analysis applications, but not for helping learn classifiers. A large number of synthetic 3D meshes
in \cite{ZX:12} were created by a series of mesh editing steps including
subdivision, simplification, smooth, adding noise and Poisson reconstruction,
in order to automatically evaluate the subjective visual quality of
a 3D object. Recently, to circumvent the point labeling difficulty
in a building roof classification problem using LiDAR point cloud,
\cite{ZX:14} explicitly indicated semantic roof points on synthetically
created roof point clouds and compute point features from the synthetic
point clouds. Techniques such as point cloud resampling, size normalization
and mesh erosion are employed to reduce the differences between real
roof and synthetic ones in data space. 

\textbf{Generating synthetic data. } Previous method generate synthetic
data in data space using tools including geometrical transformation
and degradation models: In \cite{VT:03}\cite{VT:04}, to help off-line
recognition of handwritten text, a perturbation model combined with
morphological operation is applied to real data. They showed that
when a moderate transformation is added to the real data, the resulting
synthetic training set boost the performance. To enhance the quality
of degraded document, in \cite{BG:08} degradation models such as
brightness degradation, blurring degradation, noise degradation and
texture-blending degradation were used to create a training dataset
for a handwritten text recognition problem. The synthetic minority
oversampling technique (SMOTE) \cite{CNV:02} and its variants \cite{HH:05}\cite{HH:08}
are also powerful methods that have shown many success in various applications. 
However, these previous methods are relatively limited to one particular type of dataset, whilst 
we propose a more general  methodology of generating synthetic data in this paper. 
We show that our methodology can be used both for SRC and handwritten digits dataset.

\textbf{Transfer Learning} aims to extract the knowledge from one
or more source tasks and applies the knowledge to a target task. Transfer
learning has been found helpful in many real world problems, such
as in sentiment classification \cite{Blitzer07Biographies}, web page
classification\textcolor{red}{{} }\cite{Sarinnapakorn:2007:CST:1313047.1313197}
and zero-shot classification of image and video data \cite{lampert13AwAPAMI,lampert2009zeroshot_dat,yanweiPAMIlatentattrib,yao2011action_part,yanweiembedding,yanweiBMVC,RohrbachCVPR12,rohrbach2010semantic_transfer,RichardNIPS13}.
  Transfer learning is categorized to three classes \cite{pan2009transfer_survey}:
inductive transfer learning, transductive transfer learning and unsupervised
transfer learning. The work in this paper falls into a framework of
domain adaptation \cite{Ben-David:2010:TLD:1745449.1745461,Weinberger:2009:FHL:1553374.1553516}
in the transductive transfer learning. Nonetheless, different from
previous domain adaptation tasks of different source and target domains,
the synthetic gap is caused by the shifted feature distribution of
synthetic data from real data. To solve this problem, our MCAE is
developed from the idea of autoencoder.

\textbf{Autoencoder} is one type of neural network and its output
vectors have the same dimensionality as the input vectors \cite{vincent2008ICML}.
The hidden representation obtained by training a sparse autoencoder
followed by a parameters fine tuning is useful in pre-training a deeper
neural network. Recently autoencoder with its different variants \cite{MarginalizedDenoisingAutoencoders2012ICML,Glorot11domainadaptation}
also exhibit the success in learning and transferring sharing knowledge
among data source from different domains \cite{BP:12,BY:12,DJ:13},
thus benefit other machine learning tasks.

\section{Multichannel Autoencoder (MCAE) }

\begin{figure}[ht]
\centerline{ %
\begin{tabular}{cc}
\resizebox{0.21\textwidth}{!}{\rotatebox{0}{ \includegraphics{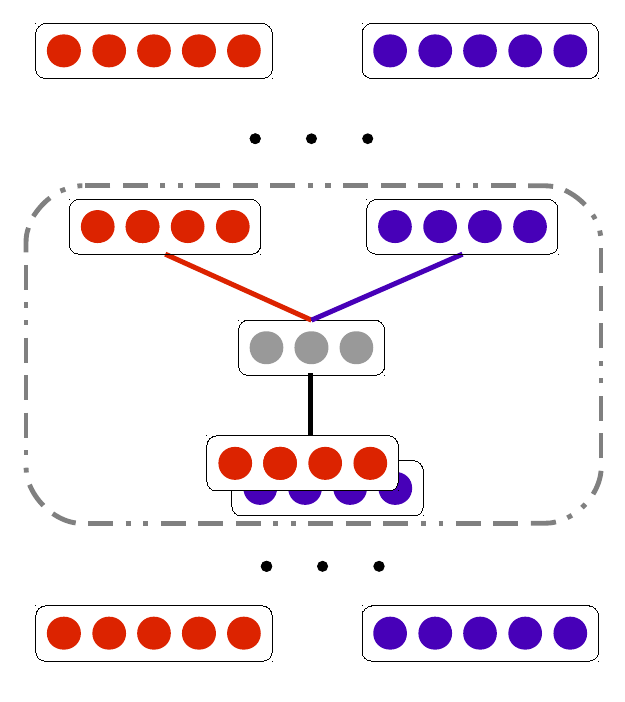}}}  & \resizebox{0.24\textwidth}{!}{\rotatebox{0}{ \includegraphics{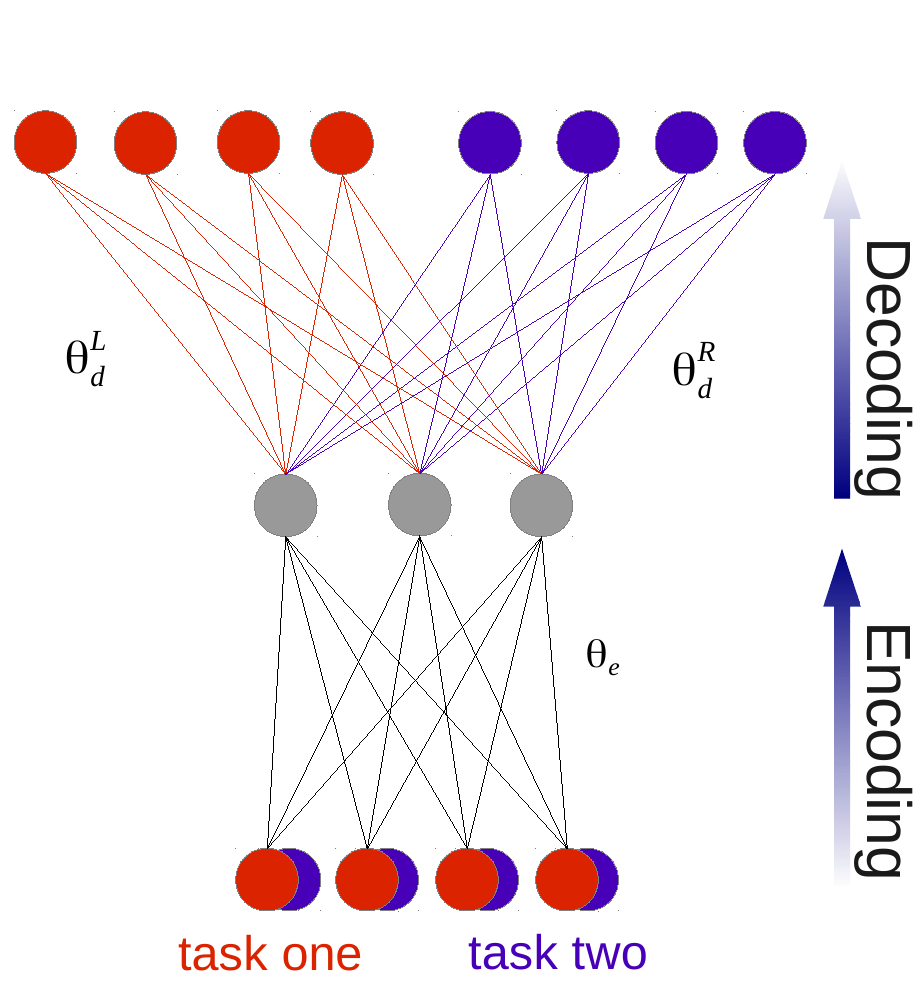}}} \tabularnewline
(a)  & (b) \tabularnewline
\end{tabular}} \protect\protect\caption{(a) Illustration of the proposed MCAE model in a stacked autoencoder
structure, where black edge between two layers are linked to and shared
by two tasks, red and blue links are separately connected to left
and right task respectively. (b) A zoom in structure of MCAE. }

\label{fig: YAE-structure} 
\end{figure}

In this section, we introduce the MCAE model as illustrated in Fig.
\ref{fig: YAE-structure}. It can (1) bridge synthetic gap by minimizing
the discrepancy between real and synthetic data; and (2) preserve
and emphasize the potential useful patterns existed in both real and
synthetic data in order to generate the better feature representations
used for learning classifiers. 
 
Essentially, synthetic and real data should have similar patterns, a natural
idea of bridging synthetic gap is to learning a mapping from the synthetic data to the real data using an autoencoder, and vice versa. MCAE, hence, provides a more flexible way to learn this mapping due to the specific structure of the MCAE. There are two channels in MCAE, left one and right one. Each channel basically is an SAE, however, two channels share the same hidden layer. With this structure, MCAE basically learns two tasks in the same time. By setting different types of input and out data such as the one in denoising autoencoder \cite{VP:10}, MCAE is capable for many applications. In our work, to bridge the gap between synthetic data and real data, we set the task in left channel as one that takes synthetic data as input and real data as \textit{reconstruction target}, while the task in right channel use real data in both input and \textit{reconstrution target}. This configuration actually is essentially meaningful that by keeping the \textit{reconstruction target} identical in two channels, MCAE attempts to transform inputs in two channels towards the same target, thus minimize the discrepancy between two input dataset which are synthetic data and real data in our work.

%

\subsection{Problem setup }

Our MCAE is built on the sparse autoencoder (SAE). A basic autoencoder
is a fully connected neural network with one hidden layer and can
be decomposed into two parts: an encoding and a decoding process.
Assume an input dataset with $n$ instances $X=\{x_{i}\}_{i=1}^{n}$
where $x_{i}\in{\mathbb{R}^{m}}$ and $m$ is the dimension of each
instance. Encoding typically transforms input data to hidden layer
representation using an affine mapping squashed by a sigmoid function:

\begin{equation}
h_{e}(x_{i})=f(W_{e}x_{i}+b_{e})\label{eq:encoder}
\end{equation}
where $f(\cdot)$ is a sigmoid function and $\theta_{e}=\{W_{e},b_{e}\}$,
$W_{e}\in\mathbb{R}^{k\times m},b_{e}\in\mathbb{R}^{k}$ is a set
of unknown parameters in encoding with $k$ nodes in hidden layer.

While in decoding, with parameters $\theta_{d}=\{W_{d},b_{d}\}$,
$W_{d}\in\mathbb{R}^{m\times k},b_{d}\in\mathbb{R}^{m}$, autoencoder
attempts to reconstruct the input data at the output layer by imposing
another affine mapping followed by nonlinearity to hidden representation
$h_{e}(x_{i})$:

\begin{equation}
h_{d}(x_{i})=f(W_{d}h_{e}(x_{i})+b_{d})\label{eq:decoding}
\end{equation}

In above equation $h_{d}(x_{i})$ is viewed as a reconstruction of
input $x_{i}$. Normally, we impose $h_{d}(x_{i})\approx x_{i}$.
Here $x_{i}$ play a role of \textit{reconstruction target} in this
expression and we use notation $\langle\mathfrak{i}\text{:}X_{i},\:\mathfrak{t}\text{:}X_{i}\rangle$
to denote the configuration of input data short for $\mathfrak{i}$
and \textit{reconstruction target} short for $\mathfrak{t}$ in an
autoencoder. $X_{s}$ and $X_{r}$ indicate synthetic and real data
respectively. By minimizing the reconstruction errors of all data
instances, we have following objective function:

\begin{equation}
J(\theta_{e},\theta_{d})=\frac{1}{n}\sum_{i=1}^{n}(h_{d}(x_{i})-x_{i})^{2}+\lambda W\label{eq:reconstruction_error_basic}
\end{equation}
where $W=({\sum W_{e}^{2}+\sum W_{d}^{2}})/2$ is a weight decay term
added to improve generalization of the autoencoder and $\lambda$
leverages the importance of this term.

To avoid learning identity mapping in autoencoder, a regularization
term $\Theta=\sum_{i=1}^{k}{\delta\text{log}\frac{\delta}{\hat{\delta}_{i}}+(1-\delta)\text{log}\frac{1-\delta}{1-\hat{\delta}_{i}}}$
that penalizes over-activation of the nodes in the hidden layer is
added. $\delta$ is a sparsity parameter and is set by users and $\hat{\delta}_{i}=\frac{1}{k}\sum_{i=1}^{k}h_{e}(x_{i})$.

\begin{equation}
J(\theta_{e},\theta_{d})=\frac{1}{n}\sum_{i=1}^{n}(\hat{x}_{i}-x_{i})^{2}+\lambda W+\rho\Theta\label{Equ: SparsityObjective}
\end{equation}
$\rho$ controls sparsity of representation in hidden layer.

Note that directly applying sparse autoencoder to our problem does
not work well. For example, we can train an autoencoder purely
by placing synthetic data in input layer and real data in output layer
denoted as $\langle\mathfrak{i}\text{:}X_{s},\:\mathfrak{t}\text{:}X_{r}\rangle$
which however can not bridge the synthetic gap in our problem. Such
way of reconstruction is only to complement the missing information
in synthetic data from real data but not vice versa%
\footnote{Please refer to supplemenatry maerial for the validation%
}. 

A better representation should be reconstructed by using the information
from both real and synthetic data simultaneously. Specifically, we
aim at two tasks: one is $\langle\mathfrak{i}\text{:}X_{s},\:\mathfrak{t}\text{:}X_{r}\rangle^L$
which reconstructs synthetic data towards real data, and the other
one is $\langle\mathfrak{i}\text{:}X_{r},\:\mathfrak{t}\text{:}X_{r}\rangle^R$
which uses identical real data for input and \textit{reconstruction
target}, where $\langle\cdot\rangle^{L}$ and $\langle\cdot\rangle^{R}$
indicate the left and right channel of MCAE.

\subsection{MCAE model}

We propose a multichannel autoencoder that uses a balance regularization
to leverage the learning between two tasks, i.e. $\langle\mathfrak{i}\text{:}X_{s},\:\mathfrak{t}\text{:}X_{r}\rangle^L$
and $\langle\mathfrak{i}\text{:}X_{r},\:\mathfrak{t}\text{:}X_{r}\rangle^R$.
The structure of this new autoencoder is shown in Fig. \ref{fig: YAE-structure}.
In this new structure, tasks of two channels will share the same parameters
$\theta_{e}$ in encoding process which will enforce autoencoder to
reconstruct common structure in both tasks. However, in decoding process,
we divide autoencoder to two separate channels that two tasks will
have their own parameters $\theta_{d}^{L}$ and $\theta_{d}^{R}$.
Dividing autoencoder to two channels at decoding layer enable a more
flexible control between the two tasks. Thus autoencoder better leverage
the common knowledge from the two tasks. 

With two channels in the MCAE, we target to minimize the reconstruction
error of two tasks together while taking into account the balance
between two channels. The new objective function of the MCAE is given
in the following:

\begin{equation}
E=J^{L}(\theta_{e},\theta_{d}^{L})+J^{R}(\theta_{e},\theta_{d}^{R})+\gamma\Psi\label{Equ: YAE-objective}
\end{equation}
where 
\begin{equation}
\Psi=\frac{1}{2}(J^{L}(\theta_{e},\theta_{d}^{L})-J^{R}(\theta_{e},\theta_{d}^{R}))^{2}
\end{equation}
is a regularization added to balance the learning rate between two
channels. This regularization will have two effects on the MCAE. First,
$\Psi$ accelerates the speed of optimizing Eq. \ref{Equ: YAE-objective},
since minimizing $\Psi$ requires both $J^{L}(\theta_{e},\theta_{d}^{L})$
and $J^{R}(\theta_{e},\theta_{d}^{R})$ are small which in turn cause
$E$ decreases faster. Second, $\Psi$ penalizes a situation more
when difference of learning error between two channels are large,
so as to avoid imbalanced learning between two channels.

The minimization of Eq. \ref{Equ: YAE-objective} is achieved by back propagation
and stochastic gradient descent using Quasi-Newton method. Since the
regularization term is added to leverage the balance of different
tasks, we have to compute the gradient of parameters $\theta_{e}$
and $\theta_{d}^{L},\theta_{d}^{R}$ in MCAE. Please refer to the
supplementary material for the detailed computation of gradients.

\subsection{The advantages of MCAE over the alternative Configurations}

Our MCAE enforces autoencoder to learn useful class patterns from
the two tasks simultaneously. Thus it helps with capturing a common
structure of synthetic and real images. Another alternative way is
to concatenate the input and target of the two tasks $\langle\mathfrak{i}\text{:}X_{s}X_{r},\:\mathfrak{t}\text{:}X_{r}X_{r}\rangle$
for autoencoder. We annotate the usage of this autoencoder as Concatenate-Input
Autoencoder (CIAE), since this autoencoder learns concatenated tasks at
the same time. Such configurations however may
result in an unbalanced optimization for these two tasks: the optimization
process of one task will take over the process of the other one. It
results in a biased reconstructed hidden layer of the autoencoder
and thus a limited classification performance. Our experiments also
validate this point in Sec. \ref{Sec: Results}.

\section{Generating Synthetic Data}

It is an important and yet less studied topic of how to generate synthetic
data. This section discusses the methodology of generating synthetic data
used in our experiments. Such synthetic data have some similarities
and differences with the augmented data used in deep learning e.g.
\cite{KrizhevskySH12}. Both of synthetic data and augmented data
aim at improving the generalisation capacity of classifiers. Nevertheless,
the methodology of generating synthetic data brings more deformed
patterns than the simply label-preserving transformations used in
data augmentation. 

Synthetic data are created to highlight the potential useful pattern
existed in real images. We have two stages of generating synthetic
data. In the first stage, for each real data used to train MCAE, a
synthetic version that best matching appearance of the real data is
generated; thus pairs of corresponding real and synthetic data can
be used to train the MCAE. In the second stage, more synthetic data
could be derived using synthetic data generated in the first stage
by both interpolation and extrapolation. To distinguish the set of
synthetic data used in these two stages, we use abbreviation \textit{Syn
I} and \textit{Syn II} to represent them respectively.

In the proposed approach, the synthetic data are represented as a
parametric model of a set of control points and edges associated to
these points in the images. From the control points, the synthetic
images could be generated to simulate the real images in terms of
having the same structure or a similar appearance. Initially, the control
points are selected from a synthetic prototype that generalize all images in
the same class. Then the locations of the control points are iteratively
optimized until convergence in order to minimize the distance between
synthetic images generated by control points and the real image. We
annotate the control points and edges associated to them as $\textbf{S}=\{\textbf{P},\textbf{E}\}$,
where $\textbf{P}=\{p_{i}\}_{i=1}^{n}$ is the set of the control points, and
$\textbf{E}=\{(p_{i},p_{j})\},1\leq i,j\leq n$ is the set of edges
connecting control points. A generalized algorithm of getting the
best matching synthetic image is provided in Algorithm \ref{Alg: MatchingSynthetic}.

\begin{algorithm}[htb]  
\small
 	\caption{ Get Matching Synthetic Image.}   
	\label{Alg: MatchingSynthetic}    	
\begin{algorithmic}[1]  	\REQUIRE ~~\\ 		
$\bullet$ A real image $U$. \\ 	
	$\bullet$ A set of control points 
$\textbf{S}=\{\textbf{P}, \textbf{E}\}$ 
with all control points $p_i\in \textbf{P}$ set to their initial positions.\\ 	
	$\bullet$ A prototype image $V$ generated using the initial $\textbf{S}$.  \\ 	
	\WHILE{\textbf{S} is not converged} 	
		\STATE \textbf{S} = OptimizeControlPoints($U, V, \textbf{S}$). 		
	\STATE Generate $V$ using $\textbf{S}$. 		
\ENDWHILE 
		\STATE Generate synthetic image $I$ using \textbf{S}. 	
	\RETURN $I$. 	
\end{algorithmic} 
 \end{algorithm}  

The synthetic prototype could be manually design or learning from given data in our work given different tasks. We will show how these two methods produce synthetic data in following two sections respectively.


\subsection{Explicitly Design of the Synthetic Prototype}
The generation of the synthetic prototype and control points in this scenario is inspired by the approach proposed by
Zhang \textit{et al.}  \cite{ZX:14}. In their work, given enough pre-knowledge about the 3D objects,
a synthetic prototype of 3D objects is explicitly designed and built. By adjusting the control points of the prototype, various kinds of 3D objects are generated. In this work, essentially, our data is very similar to theirs in a sense that roof images share a lot of characteristics such as ridge lines, valley lines and intersections between these lines, which make it possible to manually design the synthetic prototypes that characterize these patterns.  Based on this observation, a synthetic roof prototype could be generated
by setting the control points at the intersections of the ridge or
valley lines and drawing segments connecting these control points.
\footnote{In our experiments, classification of the roof images is essentially
similar to that of \cite{ZX:14}. Our approach recognizes the style
of the roofs based on edges extracted from the roof images. For more
visualisation results, please refer to our supplementary material.%
}.

In this scenario, the OptimizeControlPoints($U,V,\textbf{S}$) function of Alg. \ref{Alg: MatchingSynthetic}
turns out to be a process that searches for optimal control point locations which results in a synthetic image minimizing the discrepancy between the real image and the synthetic image. A coordinate descent framework is employed to accelerate the search process. We summarize this method in Alg. \ref{Alg: OCP1}.

\begin{algorithm}[htb]  
\small
 	\caption{ OptimizeControlPoints($U, V, \textbf{S}$) Case 1}   
	\label{Alg: OCP1}    
	\begin{algorithmic}[1]  		\REQUIRE ~~\\ 
		$\bullet$ A real image $U$. \\ 
		$\bullet$ A prototype of the synthetic image $\textbf{S}=\{\textbf{P}, \textbf{E}\}$.\\ 	
	$\bullet$ A synthetic image $V$ generated using $\textbf{S}$.  \\ 	
	\FOR{$p_i\in \textbf{P}, 1\leq i \leq n$} 		\STATE Update $\textbf{S}$ by moving $p_i$ by one unit. 	
	\STATE Generate $V$ using \textbf{S}. 	
	\IF {$\textbf{S}$ does not reduce Dist($U, V$)} 
			\STATE Cancel the last move of $p_i$. 	
		\STATE Generate $V$ using \textbf{S}. 	
	\ENDIF 	
	\ENDFOR 	
	\RETURN \textbf{S}. 	
\end{algorithmic}   
\end{algorithm}

\subsection{Learning Synthetic Prototype from Data }

%
In hand written digit dataset used in this work, we learn a synthetic prototype from given data. 
A digit prototype is generated for all images with the same digit. Congealing
algorithm proposed in \cite{ME:00} is employed in this step to produce the synthetic prototypes for digits. In congealing,
the project transformations are applied to images to minimize a joint
entropy. Thus the prototype is considered to be an average image of all
images after congealing.

Then control points are evenly sampled from the boundary detected
from the prototype image. The control points needs to be mapped
to each digit image in order to generate a synthetic image. To find this mapping we implement an
approach that migrates the control points from the prototype images to destination image.

This point migration algorithm is based on a series of intermediate
images generated in between synthetic prototype and destination image. To
generate the intermediate images, we binarize all the images and the
distance transformed images\cite{GB:86} of the synthetic prototype
and the real image are generated. Given the number of steps, an intermediate
image then is generated as a binarized image of linear interpolation
between two distance transformed images. In each
step, the control points are snapped to the closest boundary pixels
of the intermediate image. The algorithm of OptimizeControlPoints($U,V,\textbf{S}$)
in this situation is given in Algorithm \ref{Alg: OCP2}, we fix the
number of steps to $10$ in this algorithm.

\begin{figure}[ht]
\centerline{ %
\begin{tabular}{ccccccc}
\resizebox{0.045\textwidth}{!}{\rotatebox{0}{ \includegraphics{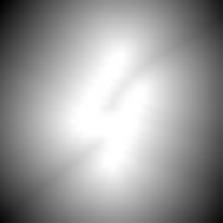}}}  
& 
\resizebox{0.045\textwidth}{!}{\rotatebox{0}{ \includegraphics{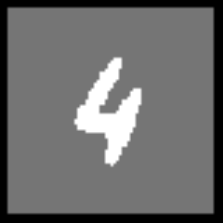}}}  
& 
\resizebox{0.045\textwidth}{!}{\rotatebox{0}{ \includegraphics{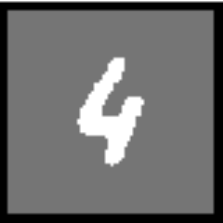}}}  
& 
\resizebox{0.045\textwidth}{!}{\rotatebox{0}{ \includegraphics{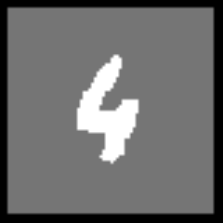}}}  
& 
\resizebox{0.045\textwidth}{!}{\rotatebox{0}{ \includegraphics{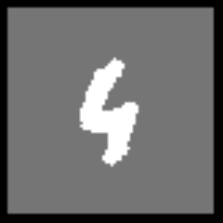}}}  
& 
\resizebox{0.045\textwidth}{!}{\rotatebox{0}{ \includegraphics{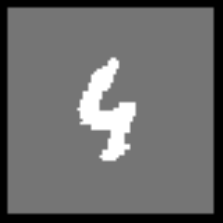}}} 
& 
\resizebox{0.045\textwidth}{!}{\rotatebox{0}{ \includegraphics{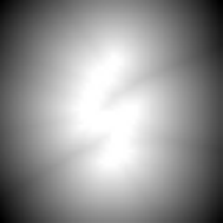}}} \tabularnewline
\end{tabular}} \caption{Illustrations of the migration of control points and intermediate
synthetic images generated using control points in each step. The
distance transform images of the synthetic prototype and real images
are shown as the left most and right most images respectively.}

\label{fig: roofexamples} 
\end{figure}

\begin{algorithm}[htb]    
\small
	\caption{ OptimizeControlPoints($U, V, \textbf{S}$) Case 2}  
 	\label{Alg: OCP2}    
	\begin{algorithmic}[1]  		\REQUIRE ~~\\ 	
	$\bullet$ A real image $U$. \\ 	
	$\bullet$ A prototype of the synthetic image $\textbf{S}=\{\textbf{P}, \textbf{E}\}$.\\ 	
	$\bullet$ A synthetic image $V$. 
\\ 	
	\STATE $steps=10$. 	
	\STATE Compute distance transform image of $U, V$ as $U', V'$. 	
	\FOR{$i=1$ to $steps$} 			
\STATE $I=(1-\frac{i}{steps})U'+\frac{i}{steps}V'$.\\
			\STATE $I$=Binarize($I$).\\ 	
		\STATE Update $\textbf{S}$ by snapping to the closest boundary pixel on $I$.\\ 	
	\ENDFOR 		\STATE Set the status of $\textbf{S}$ to be converged. 	
	\RETURN \textbf{S}. 	
\end{algorithmic} 
 \end{algorithm} 

To generate the $SynII$ dataset, we either interpolate or extrapolate
between control points of randomly choose two synthetic images from
$SynI$ dataset. The weights used in interpolation and extrapolation
is uniformly drawn from $0$ to $1$.

\section{Experiments and Results}

\label{Sec: Results} We validate the proposed MCAE dataset on several
applications in this section. This section is organised as follows.
First, in Sec 5.1, we introduce a new benchmark dataset -- Satellite
Roof Classification (SRC) dataset to vision community. This dataset
is of high quality satellite roof class labels for the satellite images. We
also briefly summarizes the handwritten digits dataset used in our
paper. We explain the experimental settings in Sec. 5.2 and discuss
the experimental results in Sec. 5.3.

\subsection{Experiment Datasets}

\subsubsection{Satellite Roof Classification (SRC) Dataset}

One particular interesting problem of learning classifiers from synthetic
data is to analyze satellite images of the Earth. Such problems generally
need very high quality (expert-level) labeled data. However, there
is no previous dataset for such research purposes. To facilitate the
study, a new benchmark Satellite Roof Classification (SRC) Dataset
is created and used in our experiments. Given a satellite image, we employ
a method described in \cite{ZX:14b} to crop roof images by
registering artificial building footprints with the satellite image.
Later, all roof images are aligned using their footprint principal
directions using a method proposed in \cite{ZQY:08} and then are
scaled to images with resolution of $128\times256$. Two experts are
invited to contribute the labels of 6 different roof styles: flat,
gable, gambrel, halfhip, hip and pyramid. Example instances of SRC
dataset are shown in Fig~\ref{fig: roofexamples}.

This dataset is of great challenges for the task of visual analysis.
First, qualities of the some satellite images are degraded because
of significant image blurring occurred when capturing the satellite
images. Second, roofs in these images are covered by various kinds
of equipments such as air conditioners chimneys and water tanks, and
most of roofs in our dataset are partially occluded by shadows cast
by trees and some other stuffs. Such covering and shadows are great
obstacles to robust visual analysis algorithms. Furthermore, the class
instances of SRC dataset are naturally extreme imbalance, since some
particular types of roofs (such as gambrel and pyramid) are far less
than the other types in the real world. Such unbalanced distributions
of data are compared in in Table \ref{Tab: datadistribution}.

\begin{figure*}[t]
\centering
\includegraphics[scale=0.32]{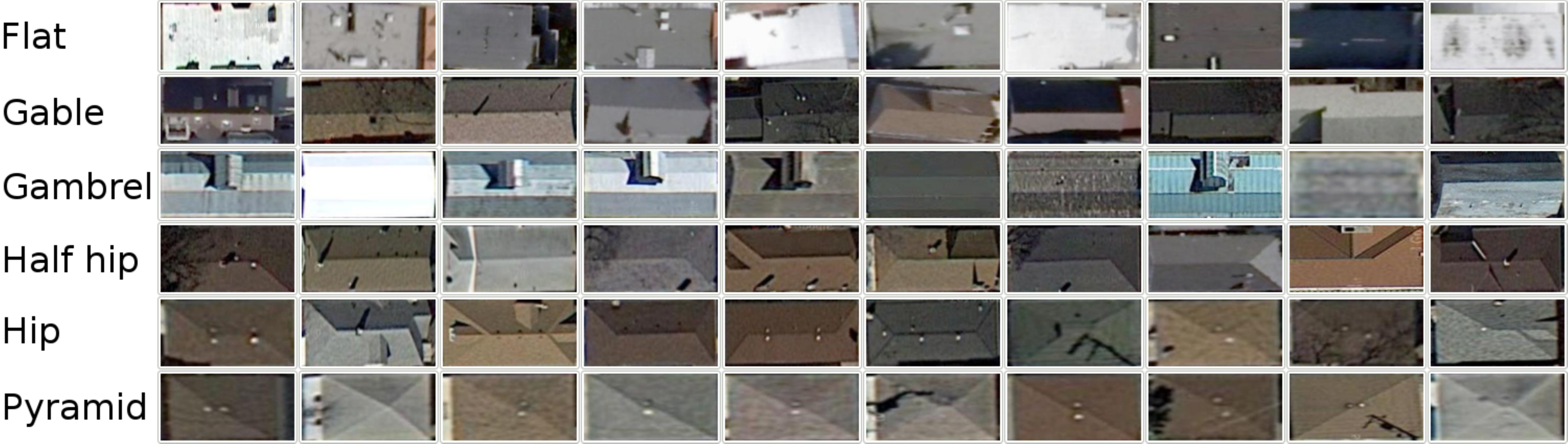}\caption{The illustration of the SRC dataset.}

\label{fig: roofexamples} 
\end{figure*}

\begin{table}[h]
\centering %
\begin{tabular}{@{}lccc@{}}
\toprule 
\textbf{Styles}  & \textbf{Training \#}  & \textbf{Testing \#}  & \textbf{Total \#} \tabularnewline
\midrule 
\textbf{Flat}  & 1232  & 1748  & 3080 \tabularnewline
\textbf{Gable}  & 1111  & 1665  & 2776 \tabularnewline
\textbf{Gambrel}  & 156  & 232  & 388 \tabularnewline
\textbf{Halfhip}  & 268  & 400  & 668 \tabularnewline
\textbf{Hip}  & 960  & 1440  & 2400 \tabularnewline
\textbf{Pyramid}  & 133  & 199  & 332 \tabularnewline
\bottomrule
\end{tabular}\protect\protect\caption{The distribution of the roof styles used in the experiments.}

\label{Tab: datadistribution} 
\end{table}

\begin{figure}[ht]
\centerline{ %
\begin{tabular}{ccc}
\resizebox{0.14\textwidth}{!}{\rotatebox{0}{ \includegraphics{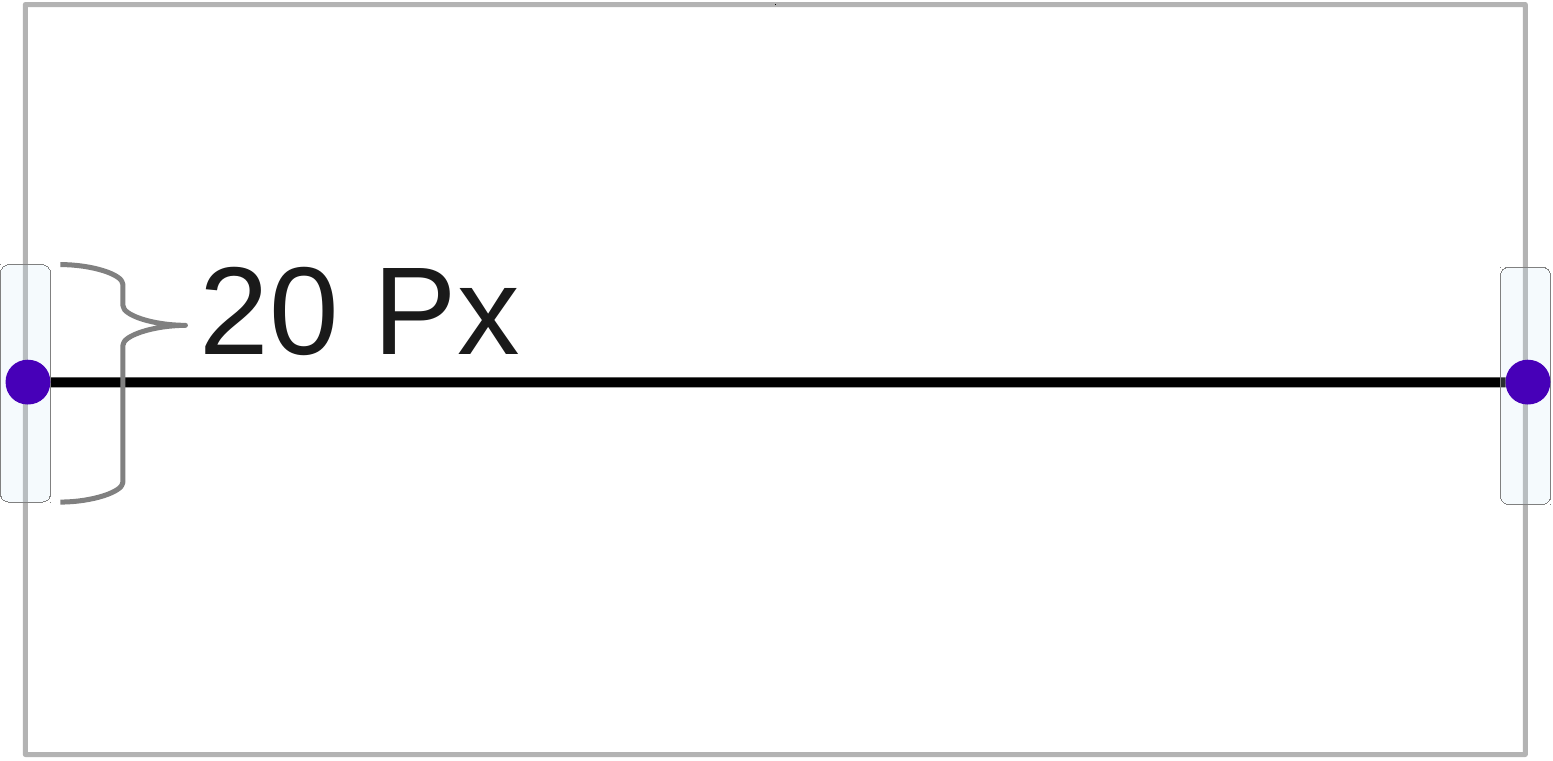}}}  & \resizebox{0.14\textwidth}{!}{\rotatebox{0}{ \includegraphics{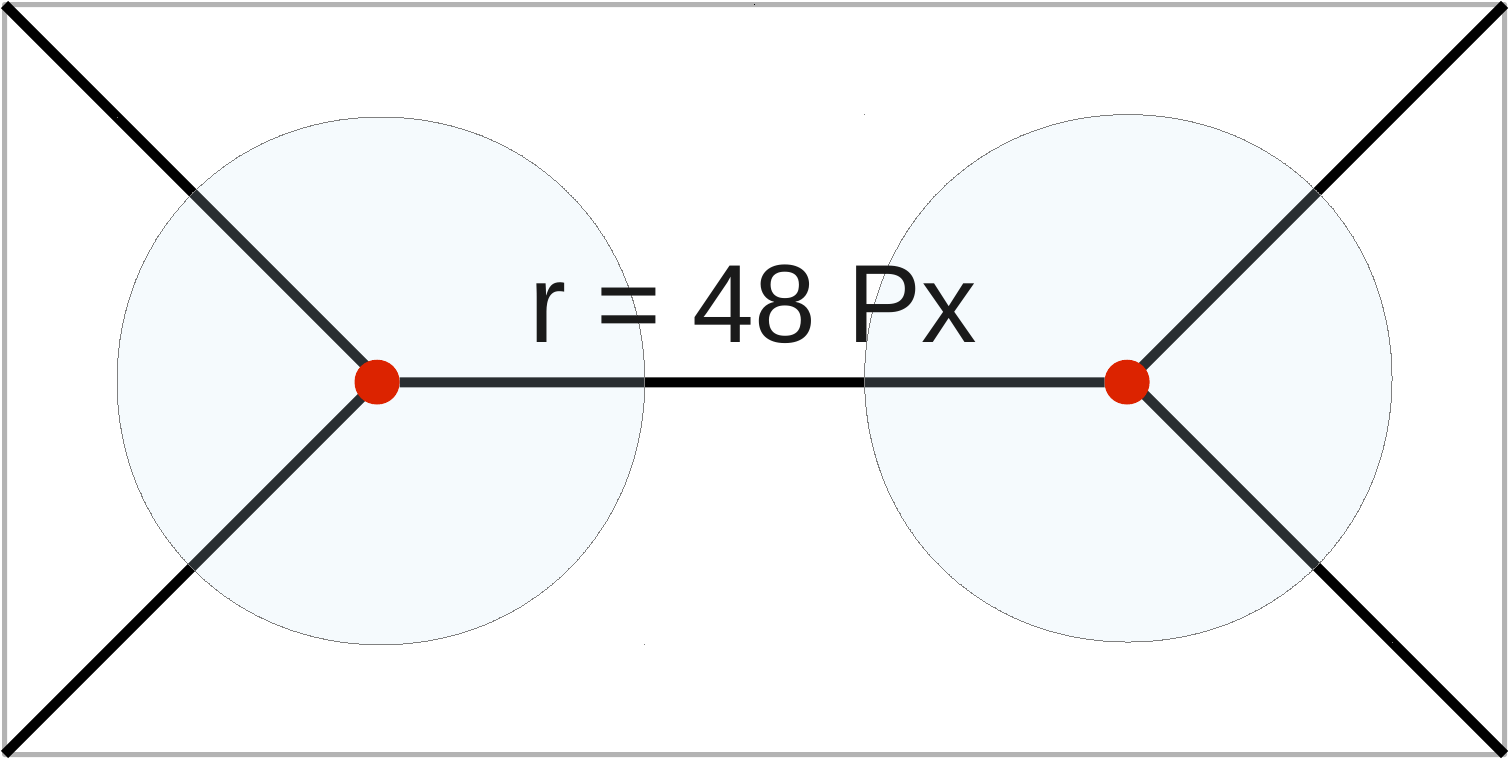}}}  & \resizebox{0.14\textwidth}{!}{\rotatebox{0}{ \includegraphics{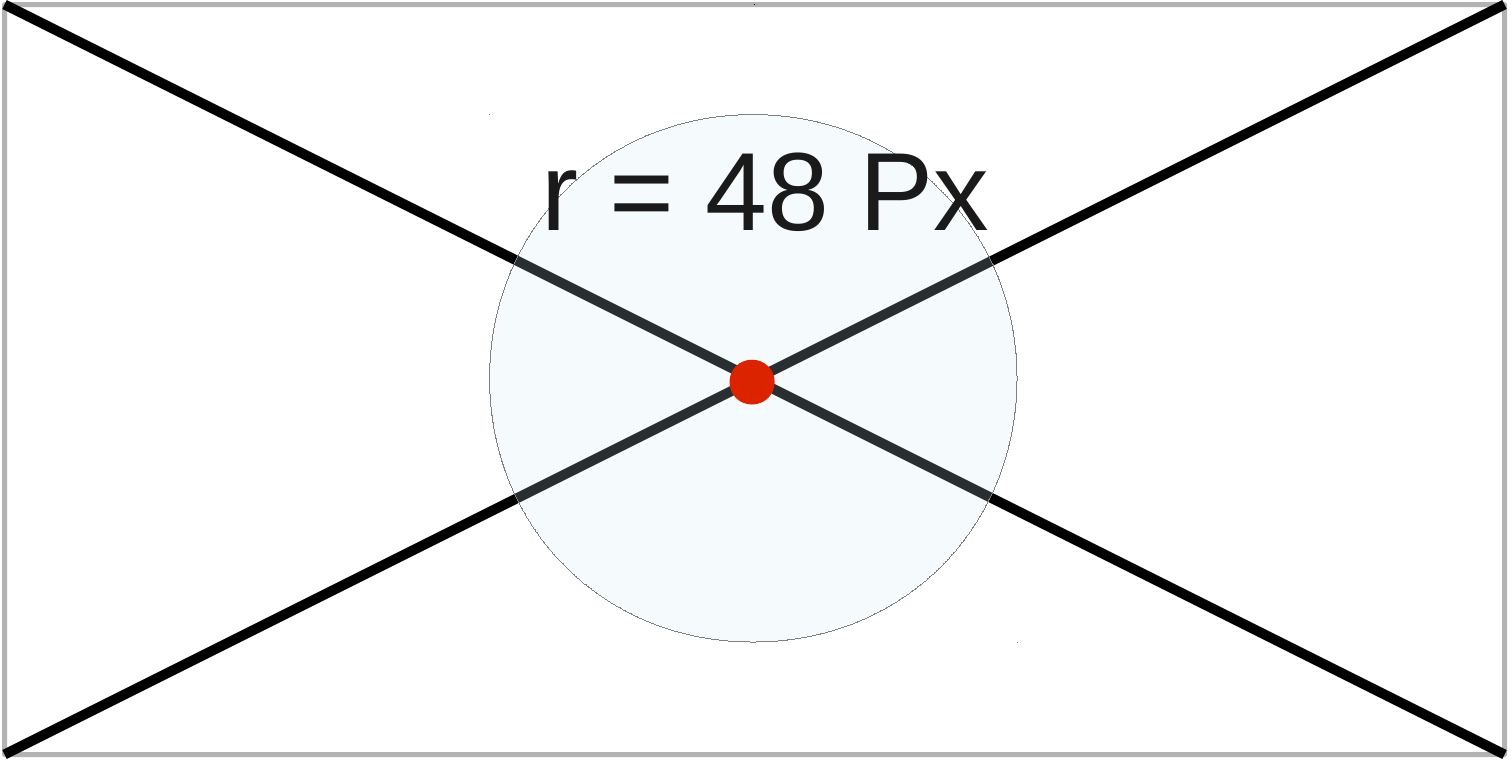}}} \tabularnewline
\end{tabular}} \protect\caption{The illustration of synthetic roof prototype we used in the process
of finding matching synthetic data for each real data. There are two
types of control points: red ones and blue ones. Blue control points
are constraint to move along the boundary only. The area of point's
movements masked using light blue. }

\label{fig: roofprototype} 
\end{figure}

Classification of the roof styles in the experiments are based on
recognizing edges detected from the roof images. We employed the adaptive
Otsu edge detection method \cite{otsu:79} to extract edges from the
roof images. We create synthetic prototype to characterize primary
ridge lines or valley lines in a certain type of roof style. Examples
of the synthetic prototypes are shown in Fig. \ref{fig: roofprototype}.
Real roof edge images and matching \textit{Syn I} images are shown
in Fig. \ref{fig: roof real and synI}. To create \textit{Syn II}
images of this dataset, $2000$ synthetic images are produced by interpolation
and extrapolation between images in \textit{Syn I} for each roof style.

\subsubsection{Handwritten Digits Dataset}

We also validate our framework on handwritten digits dataset from
UCI machine learning repository \cite{Bache+Lichman:2013} which totally
has $5620$ instances. The handwritten digits from $0$ to $9$ in
this dataset are collected from $43$ people: $30$ contributed to
the training set and the other $13$ to the test set. In the experiments,
the \textit{Syn I} data are generated using Algorithm \ref{Alg: OCP2}.
The \textit{Syn II} data of this dataset is generated using interpolation
and extrapolation as described in Sec 4.

\subsection{Experimental Settings}

We fix the configuration of MCAE as $\langle\mathfrak{i}\text{:}X_{s},\:\mathfrak{t}\text{:}X_{r}\rangle^{L}$
(left channel) and $\langle\mathfrak{i}\text{:}X_{r},\:\mathfrak{t}\text{:}X_{r}\rangle^{R}$
(right channel). Specifically, the left
channel is the reconstruction process from synthetic data to real data,
while the right channel works in the same way as a standard SAE. Our
experimental results will show that the representations learned in
such way greatly benefit the performance of classifiers we compared.

In the experiments%
\footnote{All codes (including our MCAE and creating synthetic data) will be
released once accepted.%
}, two different classifiers of utilizing learned representations from
MCAE (from Sec. 3) are compared. In the first scenario, MCAE encodes
input data to a representation (feature) in the hidden layer and a
SVM using RBF kernel is employed in this case to show the performance
of the classification. In the second scenario, MCAE takes the input
images and produces the reconstructed images at the output layer.
Features, in this case, are images, therefore can be fed to Convolutional
Neural Network (CNN) for classification. In our experiments we build
a LeNet-5 \cite{LY:98} which is originally created for digit recognition.
We show that using the same number of input data, the performance
of the CNN prefers to the data produced by the MCAE.

We summarize all evaluations and comparisons using F-1 score, which
is defined as: 
\begin{equation}
F_{1}=2\cdot\frac{\text{Precision}\cdot\text{Recall}}{\text{Precision}+\text{Recall}}
\end{equation}

\subsection{Evaluation%
\footnote{Due to the page limit, please refer to our supplementary material
for the additional experimental results.%
}}

\begin{figure*}
\centering
\begin{tabular}{cc}
\includegraphics[scale=0.45]{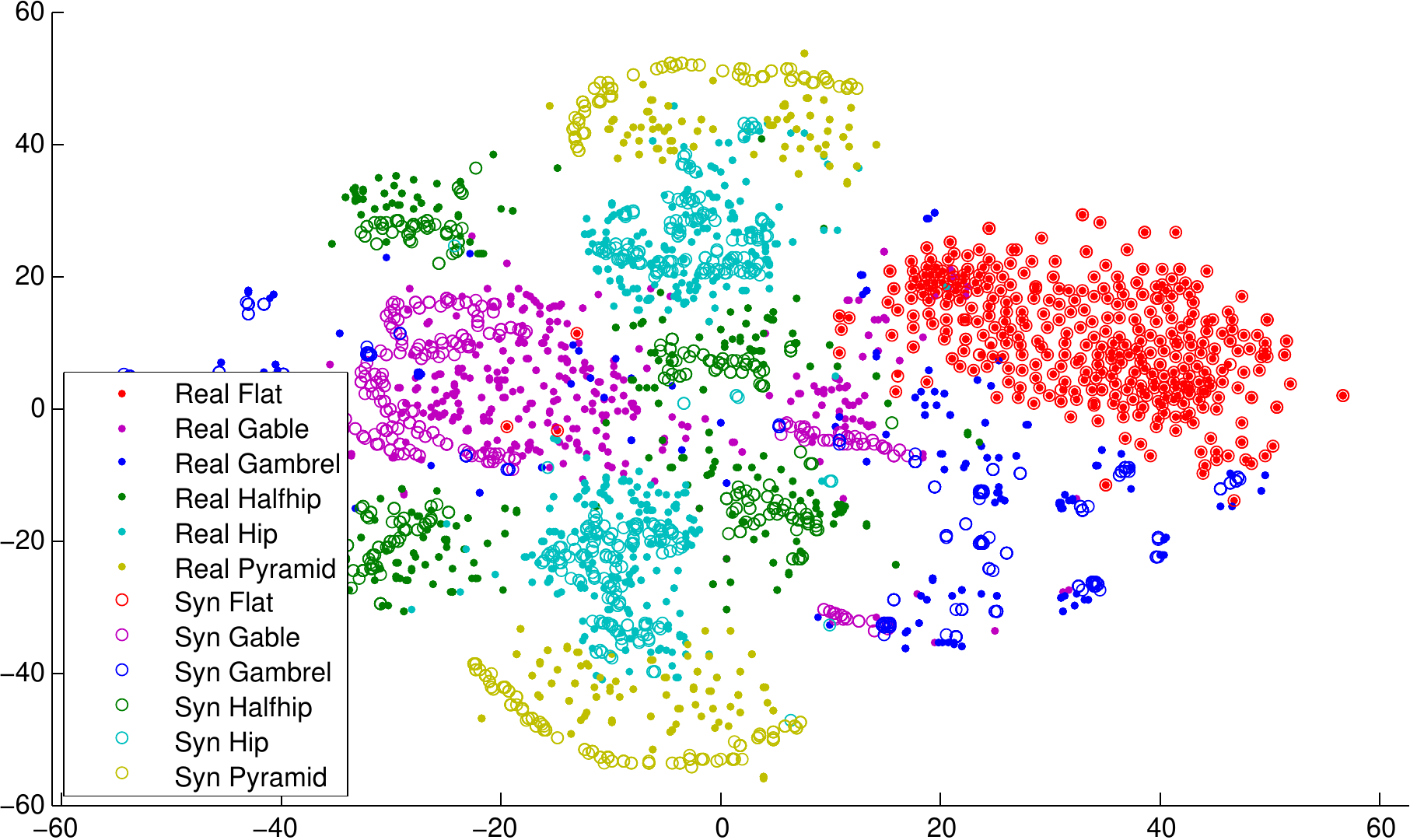} 
& 
\includegraphics[scale=0.45]{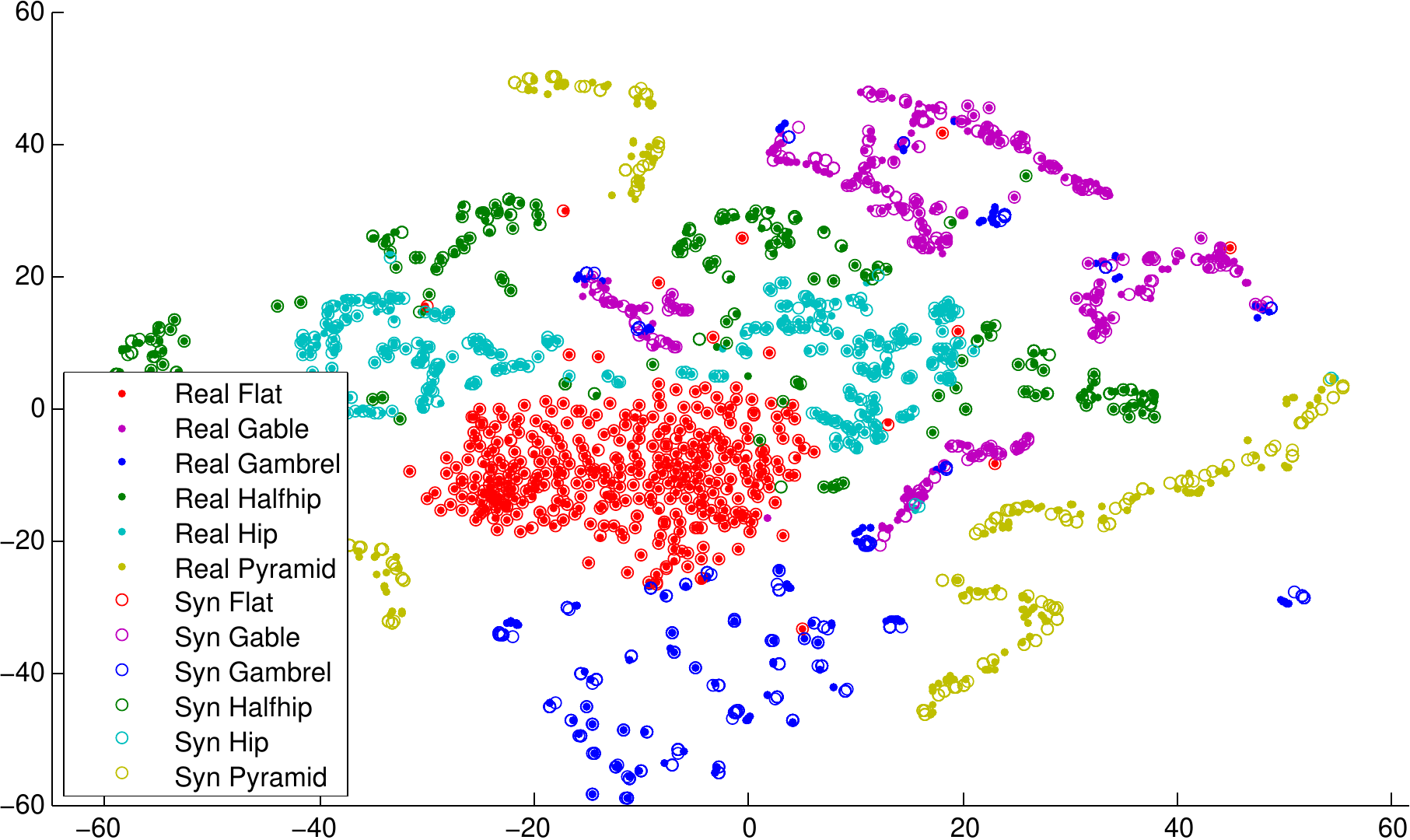}\tabularnewline
(a)  & (b) \tabularnewline
\end{tabular}

\caption{\label{fig:t-SNE-visualization-of}t-SNE \cite{tsne}visualization
of synthetic gap bridged by MCAE. (a) Data distributions of each class
of SRC dataset. For many data instances, the (circle) real and (dot
points) synthetic data are not overlapping. This is synthetic gap.
(b) Data distributions of the reconstructed images by MCAE for each
class of SRC dataset. The reconstructed images of all the real (circle)
and synthetic (dot points) are almost overlapped. It means that our
MCAE can bridge the synthetic gap.}
\end{figure*}

\textbf{\textcolor{black}{MCAE is better than CIAE and SAE. }} \quad{}To
better evaluate the performance of the proposed MCAE, we compare MCAE
with\textbf{\textcolor{red}{{} }}Concatenate-Input Autoencoder (CIAE)
\cite{multitask_autoencoder2011ICML} and Sparse Autoencoder (SAE)
\cite{VP:10}. In these experiments, we evaluate the performance on
two classifiers: a CNN using reconstructed images and SVM using encoded
hidden layer representation. We present the results of these comparisons
in Table \ref{Tab: RoofBetterAE} and Table \ref{Tab: DigitBetterAE}
for SRC and handwritten digit datasets respectively. It could be observed
from these two tables that although the performance of the CIAE is
close to MCAE, the proposed MCAE gets a better performance almost
in all the comparisons.

\begin{table}[h]
\centering %
\begin{tabular}{@{}lccc@{}}
\toprule 
 & %
\begin{tabular}{@{}c@{}}
\textbf{Data to train} \tabularnewline
\textbf{autoencoder} \tabularnewline
\end{tabular} & %
\begin{tabular}{@{}c@{}}
\textbf{CNN} \tabularnewline
\textbf{Reconstructed} \tabularnewline
\end{tabular} & %
\begin{tabular}{@{}c@{}}
\textbf{SVM} \tabularnewline
\textbf{Encoded} \tabularnewline
\end{tabular}\tabularnewline
\midrule 
\textbf{MCAE}  & %
\begin{tabular}{@{}c@{}}
$\langle\mathfrak{i}\text{:}\textit{Syn I},\:\mathfrak{t}\text{:Real}\rangle^{L}$ \tabularnewline
$\langle\mathfrak{i}\text{:}\text{Real},\:\mathfrak{t}\text{:Real}\rangle^{R}$ \tabularnewline
\end{tabular} & \textbf{0.68 }  & \textbf{0.80 }\tabularnewline
\midrule 
\textbf{CIAE}  & %
\begin{tabular}{@{}c@{}}
$\langle\mathfrak{i}\text{:}\textit{Syn I}+\text{Real}$,\tabularnewline
$\;\;\mathfrak{t}\text{:}\textit{Syn I}+\text{Real}\rangle$ \tabularnewline
\end{tabular} & 0.68  & 0.78 \tabularnewline
\midrule 
\textbf{SAE}  & $\langle\mathfrak{i}\text{:}\textit{Syn I},\:\mathfrak{t}\text{:}\textit{Syn I}\rangle$  & 0.63  & 0.59 \tabularnewline
\midrule 
\textbf{SAE}  & $\langle\mathfrak{i}\text{:Real},\:\mathfrak{t}\text{:Real}\rangle$  & 0.62  & 0.62 \tabularnewline
\bottomrule
\end{tabular}\protect\caption{F1-score of roof style classification using reconstructed images (in
CNN) and encoded image features (in SVM). Second column shows the
data used to train the autoencoder in the first column. In classification,
Real+\textit{Syn II} are used in the training of CNN and SVM. ; $\textit{Syn I}+\text{Real}$
means that we use concatenation of the Syn I and real images as the input for
the corresponding autoencoders.}

\label{Tab: RoofBetterAE} 
\end{table}

\begin{table}[h]
\centering %
\begin{tabular}{@{}lccc@{}}
\toprule 
 & %
\begin{tabular}{@{}c@{}}
\textbf{Data to train} \tabularnewline
\textbf{autoencoder} \tabularnewline
\end{tabular} & %
\begin{tabular}{@{}c@{}}
\textbf{CNN} \tabularnewline
\textbf{Reconstructed} \tabularnewline
\end{tabular} & %
\begin{tabular}{@{}c@{}}
\textbf{SVM} \tabularnewline
\textbf{Encoded} \tabularnewline
\end{tabular}\tabularnewline
\midrule 
\textbf{MCAE}  & %
\begin{tabular}{@{}c@{}}
$\langle\mathfrak{i}\text{:}\textit{Syn I},\:\mathfrak{t}\text{:Real}\rangle^{L}$ \tabularnewline
$\langle\mathfrak{i}\text{:}\text{Real},\:\mathfrak{t}\text{:Real}\rangle^{R}$ \tabularnewline
\end{tabular} & \textbf{0.98 }  & \textbf{0.96 }\tabularnewline
\midrule 
\textbf{CIAE}  & %
\begin{tabular}{@{}c@{}}
$\langle\mathfrak{i}\text{:}\textit{Syn I}+\text{Real}$,\tabularnewline
$\;\;\mathfrak{t}\text{:}\textit{Syn I}+\text{Real}\rangle$ \tabularnewline
\end{tabular} & 0.97  & 0.96 \tabularnewline
\midrule 
\textbf{SAE}  & $\langle\mathfrak{i}\text{:}\textit{Syn I},\:\mathfrak{t}\text{:}\textit{Syn I}\rangle$  & 0.94  & 0.91 \tabularnewline
\midrule 
\textbf{SAE}  & $\langle\mathfrak{i}\text{:Real},\:\mathfrak{t}\text{:Real}\rangle$  & 0.95  & 0.65 \tabularnewline
\bottomrule
\end{tabular}\protect\caption{F1-score of handwritten digit recognition.}

\label{Tab: DigitBetterAE} 
\end{table}

\textbf{Synthetic data help learning a better classifier}. \quad{}We
designed another group of experiments. In these experiments three
different configurations of data are either reconstructed and encoded
using the proposed MCAE, then used to train a CNN or a SVM in the
experiments. All results from these experiments are compared in Table
\ref{Tab: RoofBetterData} and Table \ref{Tab: DigitBetterData} respectively.
An interesting thing to notice is that in experiments, using synthetic
data can only achieve the same result as using a combination of real
and synthetic data. This result proves that the distribution of the
real data in this case is almost overlapping with the distribution
of the synthetic data.

\begin{table}[h]
\centering %
\begin{tabular}{@{}lcccc@{}}
\toprule 
 & \textbf{Feature type}  & \textbf{Real}  & \textbf{\textit{Syn II}}  & \textbf{Real+}\textbf{\textit{Syn II}} \tabularnewline
\midrule 
\textbf{CNN}  & Reconstructed  & 0.65  & 0.68  & 0.68 \tabularnewline
\textbf{SVM}  & Encoded  & 0.77  & 0.78  & 0.80 \tabularnewline
\bottomrule
\end{tabular}\protect\caption{F1-score of roof style classification by classifier (CNN and SVM)
using different set of data reconstructed of encoded using the proposed
MCAE.}

\label{Tab: RoofBetterData} 
\end{table}

\begin{table}[h]
\centering %
\begin{tabular}{@{}lcccc@{}}
\toprule 
 & \textbf{Feature type}  & \textbf{Real}  & \textbf{\textit{Syn II}}  & \textbf{Real+}\textbf{\textit{Syn II}} \tabularnewline
\midrule 
\textbf{CNN}  & Reconstructed  & 0.94  & 0.96  & 0.96 \tabularnewline
\textbf{SVM}  & Encoded  & 0.96  & 0.96  & 0.98 \tabularnewline
\bottomrule
\end{tabular}\protect\caption{F1-score of handwritten digit recognition.}

\label{Tab: DigitBetterData} 
\end{table}

\begin{figure}[ht]
\centerline{ %
\begin{tabular}{c}
\resizebox{0.45\textwidth}{!}{\rotatebox{0}{ \includegraphics{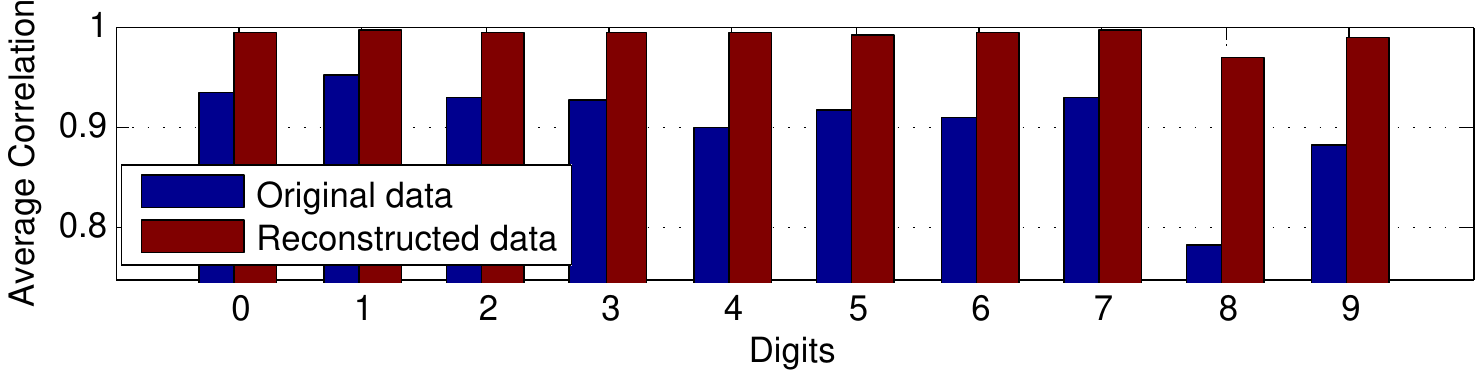}}} \tabularnewline
\tabularnewline
\resizebox{0.45\textwidth}{!}{\rotatebox{0}{ \includegraphics{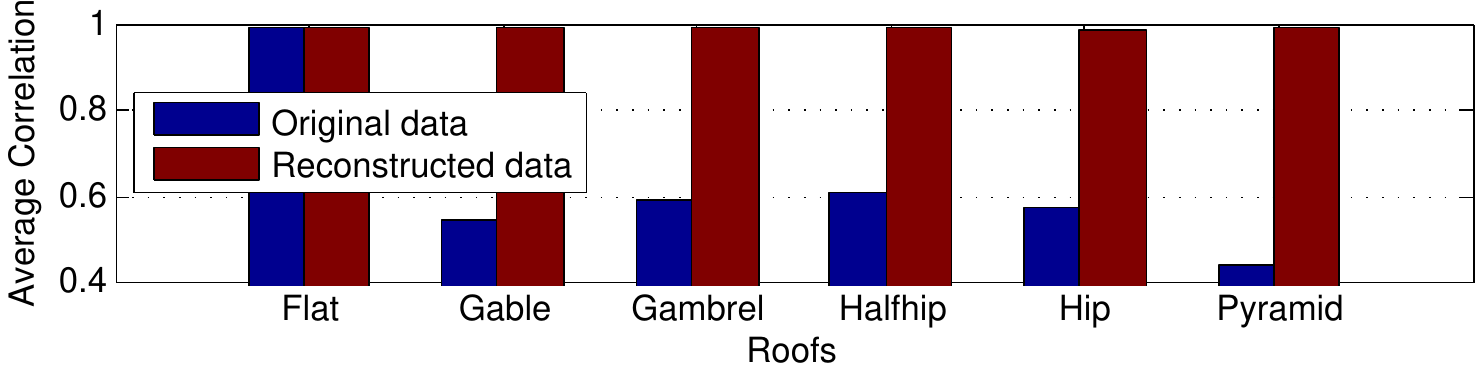}}} \tabularnewline
\end{tabular}} \protect\caption{Correlation between real and corresponding best matching \textit{Syn
I} data.}

\label{fig: correlation} 
\end{figure}

\textbf{MCAE bridges the synthetic gap. } \quad{}We compare the correlation
defined as: 
\begin{equation}
\text{Corr}=\frac{\text{Cov}(X,Y)}{\text{Var}(X)\text{Var}(Y)}
\end{equation}
between real and \textit{Syn I} data before and after being reconstructed
by the MCAE. The intention of these comparisons is to show that real
synthetic images become much more alike each other in terms of the
appearance after being reconstructed by the MCAE. The results are
shown in Fig. \ref{fig: correlation}. It is shown that our method
almost achieves $100\%$ correlation between real and \textit{Syn
I} when both data are reconstructed by the proposed MCAE. That means
the proposed MCAE bridges the synthetic gap between the real data
and the synthetic data. The results are shown in Fig.~\ref{fig:t-SNE-visualization-of}.
It intuitively shows that our MCAE can help bridge the synthetic gap
between real and synthetic data.

\section{Conclusion}

In this paper we identify the problem of synthetic gap. By solving this problem, in our experiments, we demonstrate that the synthetic data could be used to improve the performance of classifiers. To better learn classifiers from synthetic data, we have proposed a novel Multichannel autoencoder (MCAE) model. MCAE has multiple channels in its structure and is an extension from standard autoencoder. We show that MCAE not only bridges the synthetic gap between real data and synthetic data, it also jointly learns from both real and synthetic data, thus can provide more robust representation for both data. To facilitate the study on satellite image analysis, we introduce a novel benchmark dataset -- SRC as one dataset used in our experiments. The proposed method has been validated on SRC and handwritten digits datasets.

{\small{}\bibliographystyle{abbrv}
\bibliography{mybib_v2}
 } 
 
\section*{Supplementary material }

\subsection{Optimization of MCAE}

With two branches in the MCAE, we target to minimize the reconstruction
error of two tasks together while taking into account the balance
between two branches. The new objective function of the YMAE is given
in the following:

\begin{equation}
E=J^{L}(\theta_{e},\theta_{d}^{L})+J^{R}(\theta_{e},\theta_{d}^{R})+\gamma\Psi\label{Equ: YAE-objective}
\end{equation}
where 
\begin{equation}
\Psi=\frac{1}{2}(J^{L}(\theta_{e},\theta_{d}^{L})-J^{R}(\theta_{e},\theta_{d}^{R}))^{2}
\end{equation}
is a regularization added to balance the learning rate between two
branches. This regularization will have two effects on the YMAE. First,
$\Psi$ accelerates the speed of optimizing Eq. \ref{Equ: YAE-objective},
since minimizing $\Psi$ requires both $J^{L}(\theta_{e},\theta_{d}^{L})$
and $J^{R}(\theta_{e},\theta_{d}^{R})$ are small which in turn cause
$E$ decrease faster. Second, $\Psi$ penalize a situation more when
difference of learning error between two branches are large, so as
to avoid imbalanced learning between two branches.

The minimization of Eq. \ref{Equ: YAE-objective} is achieved by back
propagation and stochastic gradient descent using Quasi-Newton method.
In the MCAE, with balance regularization added to the objective, the
only difference as opposed to sparse autoencoder is the gradient computation
of unknown parameters $\theta_{e}$ and $\theta_{d}^{L},\theta_{d}^{R}$.
We clarify these differences in the following equations:

\begin{equation}
\begin{split}\nabla_{W_{e}}E= & \frac{\partial{J^{L}}}{\partial{W_{e}}}+\frac{\partial{J^{R}}}{\partial{W_{e}}}+\gamma(J^{L}-J^{R})(\frac{\partial{J^{L}}}{\partial{W_{e}}}-\frac{\partial{J^{R}}}{\partial{W_{e}}})\\
\nabla_{b_{e}}E= & \frac{\partial{J^{L}}}{\partial{b_{e}}}+\frac{\partial{J^{R}}}{\partial{b_{e}}}+\gamma(J^{L}-J^{R})(\frac{\partial{J^{L}}}{\partial{b_{e}}}-\frac{\partial{J^{R}}}{\partial{b_{e}}})
\end{split}
\end{equation}
and 
\begin{equation}
\begin{split} & \nabla_{W_{d}^{L}}E=\frac{\partial{J^{L}}}{\partial{W_{d}^{L}}}+\gamma(J^{L}-J^{R})\frac{\partial{J^{L}}}{\partial{W_{d}^{L}}}\\
 & \nabla_{b_{d}^{L}}E=\frac{\partial{J^{L}}}{\partial{b_{d}^{L}}}+\gamma(J^{L}-J^{R})\frac{\partial{J^{L}}}{\partial{b_{d}^{L}}}\\
 & \nabla_{W_{d}^{R}}E=\frac{\partial{J^{R}}}{\partial{W_{d}^{R}}}+\gamma(J^{L}-J^{R})(-\frac{\partial{J^{R}}}{\partial{W_{d}^{R}}})\\
 & \nabla_{b_{d}^{R}}E=\frac{\partial{J^{R}}}{\partial{b_{d}^{R}}}+\gamma(J^{L}-J^{R})(-\frac{\partial{J^{R}}}{\partial{b_{d}^{R}}})
\end{split}
\end{equation}

The exact form of gradients of $\theta_{e}$ and $\theta_{d}^{L},\theta_{d}^{R}$
varies according to different sparsity regularization $\Theta$ used
in the framework.

\section{Generating synthetic data}

\begin{figure}[tph]
\centering %
\begin{tabular}{c}
\resizebox{0.54\textwidth}{!}{\rotatebox{0}{ \includegraphics{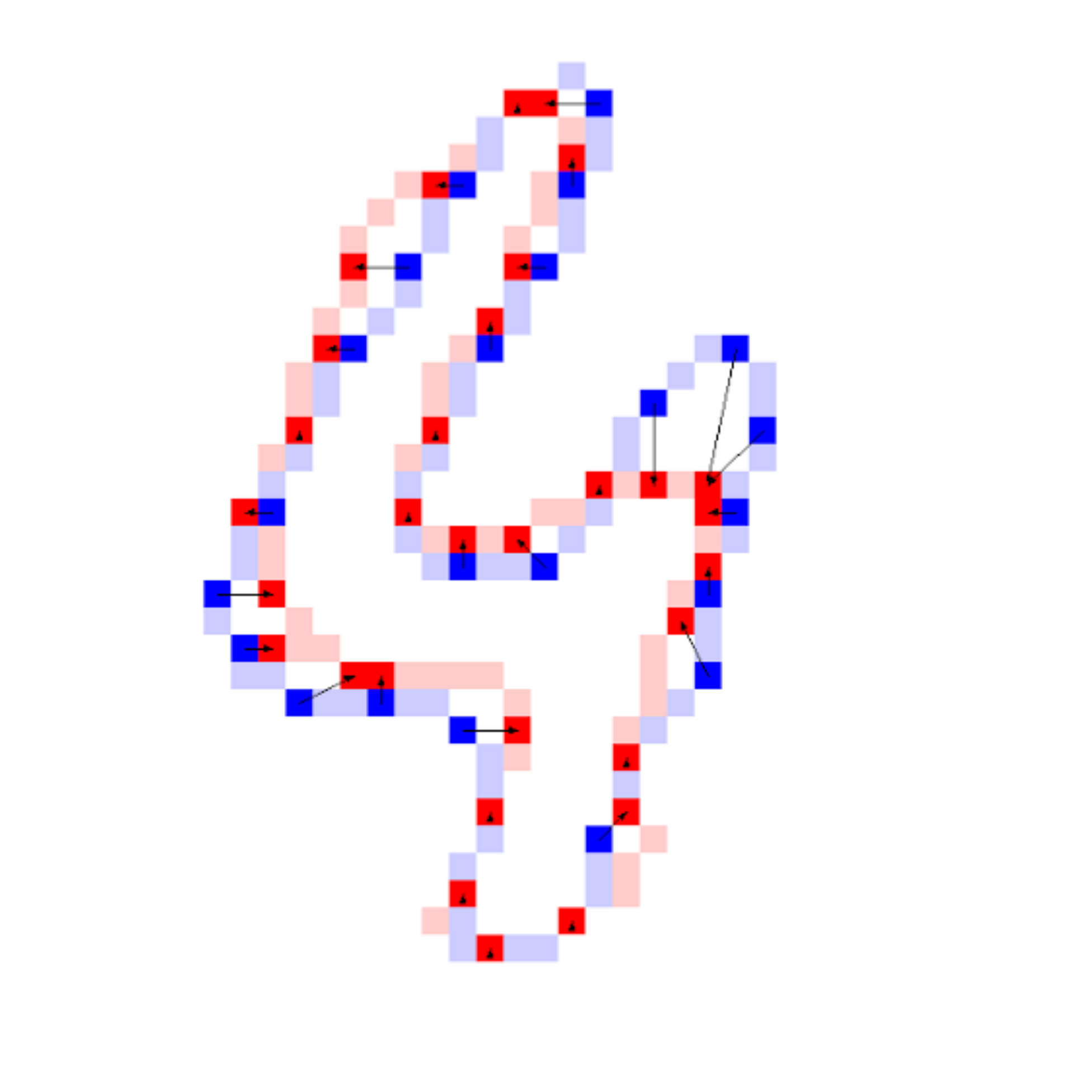}}} \tabularnewline
\end{tabular}\protect\protect\caption{An example of migration of the control points from source image (blue)
to destination image (red).}

\label{fig: migration} 
\end{figure}

An example in Fig.~\ref{fig: migration} shows how control points
are moved from source image (the one with blue boundary) to destination
image (the one with red boundary). It could be observed that most
of the control points are moved from the source image to corresponding
locations on destination image. In this step, it is not necessary
for all control points accurately move to exact corresponding location
on the destination image. Our goal is just to use these migrated control
points to generate synthetic data which will roughly mimic the real
data. Our MCAE later will rectify the difference between synthetic
data and real data.

\subsection{Further Validation for MCAE}

Note that directly applying sparse autoencoder to our problem does
not work well. For example, we can train an autoencoder purely by
placing synthetic data in input layer and real data in output layer
which however can not bridge the synthetic gap in our problem. Such
way of reconstruction is only to complement the missing information
in synthetic data from real data. On the contrary, reconstructed real data using 
such SAE will add unnecessary information and noisy patterns to reconstructed 
data. 

\begin{figure}[tph]
\centering %
\begin{tabular}{cc}
\resizebox{0.45\textwidth}{!}{\rotatebox{0}{ \includegraphics{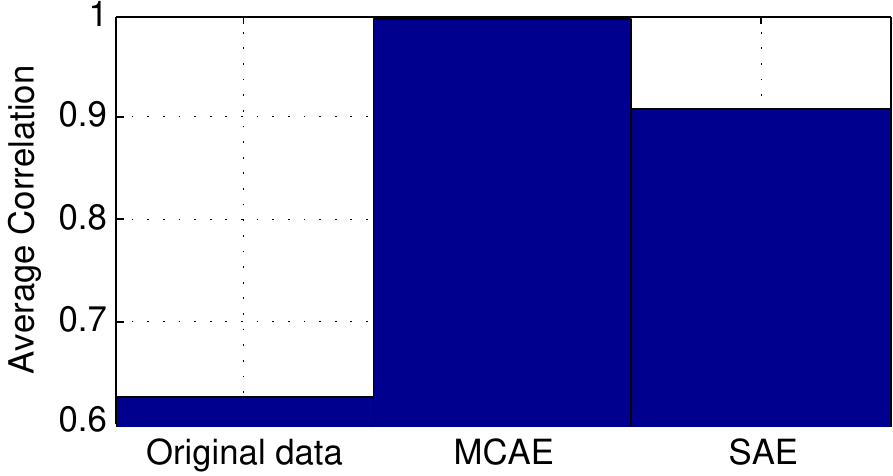}}} &
\resizebox{0.45\textwidth}{!}{\rotatebox{0}{ \includegraphics{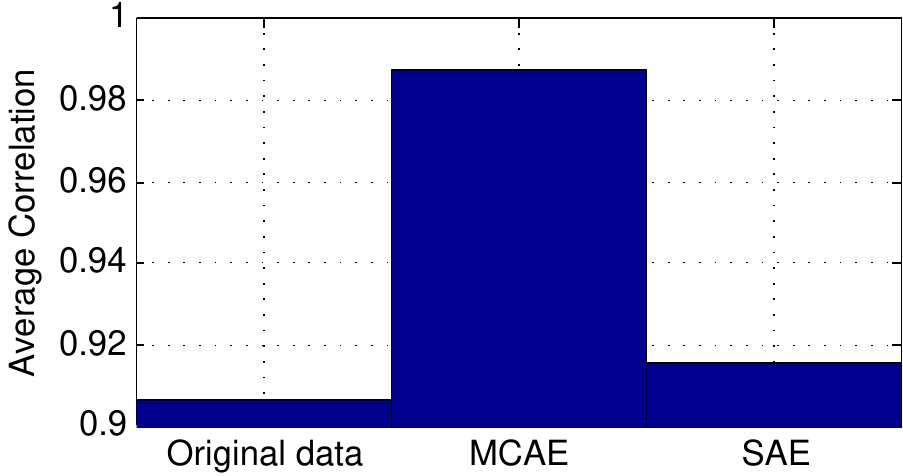}}} \\

Roof & Digit \\
\end{tabular}\protect\caption{MCAE almost perfectly bridge the synthetic gap and is much better than SAE on this job. }

\label{fig: 3bars} 
\end{figure}

To validate this point, we extend the experiments of two datasets
and show that SAE can not bridge the gap of synthetic gap. The results
are shown in Fig. \ref{fig: 3bars}. The reconstructed data of SAE
have lower average divergence than the other methods. That means,
SAE performances worse than MCAE in bridging the synthetic gap.

{\small{}{}  }

\end{document}